\documentclass[letterpaper, 10 pt, conference]{ieeeconf}  % Comment this line out if you need a4paper

\IEEEoverridecommandlockouts                              % This command is only needed if 
\usepackage{cite}
\usepackage{amsmath,amssymb,amsfonts}
\usepackage{algorithmic}
\usepackage{graphicx}
\usepackage{textcomp}
\usepackage{subcaption}
\usepackage{url}
\usepackage{xcolor}
\usepackage{multirow}
\usepackage{balance}

\title{\LARGE \bf
Why Robots Are Bad at Detecting Their Mistakes: \newline Limitations of Miscommunication Detection in Human-Robot Dialogue
}

\author{Ruben Janssens$^{1}$, Jens De Bock$^{1}$, Sofie Labat$^{2}$, Eva Verhelst$^{1}$, Veronique Hoste$^{2}$, Tony Belpaeme$^{1}$% <-this % stops a space
\thanks{This research received funding from the Flemish Government (AI Research Program 2) and from the Research Foundation Flanders (FWO Vlaanderen, 1S96324N and 1S50425N).}
\thanks{$^{1}$IDLab-AIRO, Ghent University--imec, Ghent, Belgium, contact:\newline
        {\tt\small ruben.janssens@ugent.be}}%
\thanks{$^{2}$LT3, Ghent University, Ghent, Belgium}%
}

\begin{document}

\maketitle
\thispagestyle{empty}
\pagestyle{empty}

%%%%%%%%%%%%%%%%%%%%%%%%%%%%%%%%%%%%%%%%%%%%%%%%%%%%%%%%%%%%%%%%%%%%%%%%%%%%%%%%
\begin{abstract}

Detecting miscommunication in human-robot interaction is a critical function for maintaining user engagement and trust. While humans effortlessly detect communication errors in conversations through both verbal and non-verbal cues, robots face significant challenges in interpreting non-verbal feedback, despite advances in computer vision for recognizing affective expressions. This research evaluates the effectiveness of machine learning models in detecting miscommunications in robot dialogue. Using a multi-modal dataset of 240 human-robot conversations, where four distinct types of conversational failures were systematically introduced, we assess the performance of state-of-the-art computer vision models. After each conversational turn, users provided feedback on whether they perceived an error, enabling an analysis of the models' ability to accurately detect robot mistakes. Despite using state-of-the-art models, the performance barely exceeds random chance in identifying miscommunication, while on a dataset with more expressive emotional content, they successfully identified confused states. To explore the underlying cause, we asked human raters to do the same. They could also only identify around half of the induced miscommunications, similarly to our model. These results uncover a fundamental limitation in identifying robot miscommunications in dialogue: even when users perceive the induced miscommunication as such, they often do not communicate this to their robotic conversation partner. This knowledge can shape expectations of the performance of computer vision models and can help researchers to design better human-robot conversations by deliberately eliciting feedback where needed.
%However, the models barely perform better than random chance at predicting the user’s perception. To investigate the root cause of this failure, we compare with human raters, showing that they too are only able to identify about half of the perceived mistakes. Furthermore, we apply the same computer vision model to a dataset of videos that contain much more expressive emotions, showing that the model is perfectly capable of identifying confusion.
% However, the models barely perform better than random chance at predicting the user's perception. To investigate the root cause of this failure, we apply the same computer vision model to a dataset of videos that contain much more expressive emotions, showing that the model is perfectly capable of identifying confusion. Furthermore, we also compare with human raters, showing that they too are only able to identify about half of the perceived mistakes.
% These results uncover a fundamental limitation in identifying robot mistakes in dialogue: even when users actually perceive the introduced mistakes as such, they often do not communicate this to their robotic conversation partner. This knowledge can shape expectations of the performance of computer vision models and can help researchers to better design human-robot conversations to deliberately elicit feedback where needed.
\end{abstract}

% People are not a toy problem

% \begin{IEEEkeywords}
% human-robot interaction, adaptive dialogue, robot mistakes, affect recognition
% \end{IEEEkeywords}

\section{Introduction}
% Ruben

In dialogue, individuals do more than merely interpret their interlocutors' words; they also seek feedback regarding the ongoing interaction. This process, as part of grounding the conversation, involves communicative partners signalling their understanding and agreement on the information exchanged. Through grounding, interlocutors adjust their contributions to ensure that the intended meaning has been successfully conveyed \cite{clark1996using}. This capability is
also crucial for socially interactive robots, as detecting miscommunications is key to handling the complexity and inherent imperfections in human-robot dialogue.

Detecting feedback on miscommunications is a multimodal affair: people express their feedback through speech, but also through gaze, head gestures, body pose, and in particular facial expressions \cite{axelsson2022modeling, Kontogiorgos2021ASC}.
Recent advances in computer vision, particularly through the application of Deep Learning, have significantly enhanced the performance of machine learning models for facial expression recognition \cite{li2020deep}. However, it remains unclear how these advances transfer to real-world scenarios.

\begin{figure}
    \centering
    \includegraphics[width=\columnwidth]{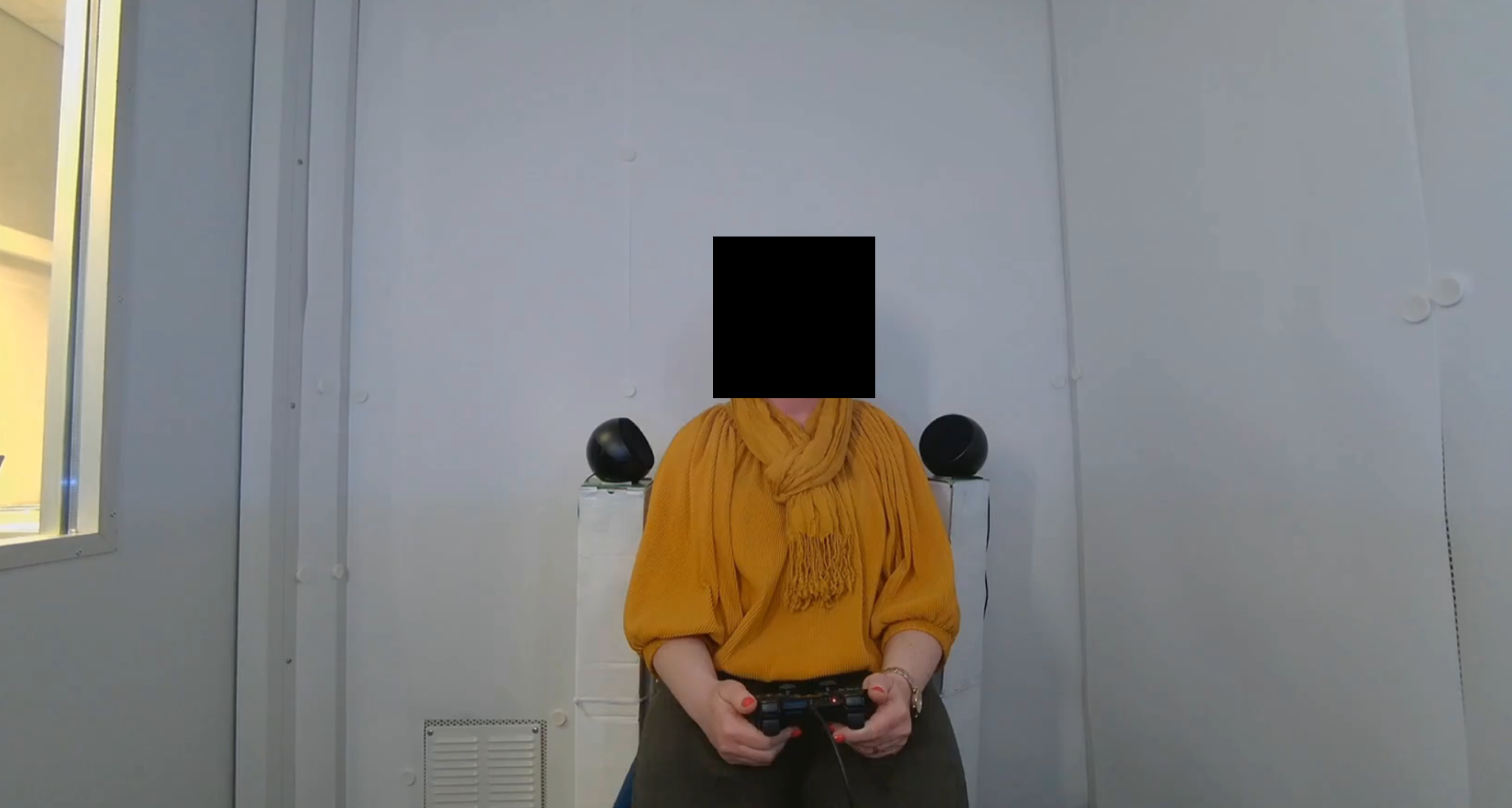}
    \caption{An anonymised example frame from the REPAIR-Corpus dataset \cite{furhat_cooking_data}, showing the user from the robot's perspective during their conversation.}
    \label{fig:example-frame}
\end{figure}

To build robotic systems capable of detecting mistakes and misunderstandings in dialogue, we set out to investigate the performance of machine learning models at detecting such communication breakdowns in real human-robot conversations. For this, we read the non-verbal signals of the human interlocutor during an interaction with a robot, and predict if and when they struggle to comprehend the robot.

Whereas previous research predominantly examines task-oriented collaborative interactions, in which humans and robots cooperate to complete a (physical) task with guidance from the robot \cite{reciperobot, stiber2023using, headandshoulder}, this study explores an educational context in which the robot assumes the role of an instructor, explaining a topic to the user while periodically posing questions to maintain engagement. Within this dialogue, four distinct types of intentional errors are introduced.

Previous studies have relied on external annotators to classify feedback as either positive or negative or solely relied on the intended effect of communication errors that were explicitly introduced \cite{axelsson2022multimodal, reciperobot, stiber2023using, headandshoulder}. In contrast, our approach seeks to determine whether users perceive the robot's speech acts as erroneous. To achieve this, we collected real-time feedback directly from users during conversations. This turn-by-turn feedback serves as the ground truth for training our machine learning models. Figure~\ref{fig:example-frame} shows an example frame from the dataset, showing the user from the perspective of the robot.% The dataset has been published on Zenodo \cite{furhat_cooking_data}.

In this paper, we describe in detail the construction of a system that detects if a user felt the robot caused a miscommunication. The system operates on videos taken from the robot's perspective, showing the torso and head of the interlocutor, as both facial expressions and head movements were identified as strong predictors of negative feedback in earlier studies \cite{axelsson2022multimodal, kontogiorgos2020behavioural}. The system contains two main components: a model to extract salient fragments from the interaction videos, and another model to classify those fragments as indicative of a misunderstanding or not.

Upon observing that the system's performance significantly underperforms relative to initial expectations, we conducted further validation by training and evaluating the model using a dataset of short videos featuring highly expressive emotional content. Additionally, we explored human performance on the same task by instructing participants to label the same videos used in the model's evaluation.
%We finally investigate the role foundation models can play in this problem, evaluating their performance at classifying verbal as well as non-verbal feedback.

%In conclusion, this study addresses the disparity between affect recognition in controlled laboratory settings and the detection of miscommunication in real-world human-robot interactions. It examines the critical distinctions in human behaviour during real-world interactions compared to the behaviour seen in existing, lab-collected datasets, providing insights into why the former task is particularly challenging.

In conclusion, this study critically examines the successes reported in previous affect recognition and robot failure detection research. We apply state-of-the-art techniques for processing facial expressions for feedback detection to a social human-robot dialogue in an educational setting and provide insights into why this task is so challenging by leveraging (1) explicit feedback directly from the user in the interaction as a ground truth and (2) investigating human performance when acting as external annotators.

% agnes: feedback is elicited (but not always), also not using facial expressions in the end, mostly using head movements using a headset

% WE HAVE TURN-BY-TURN FEEDBACK, HUMAN GROUND TRUTH: THIS IS UNIQUE

% human teachers constantly adapt

% robots (LLMs) lack the capability to detect when they made a mistake, people do this (core element of dialogue): negotiating / grounding / feedback channels (thesis/review Agnes)

% no shared goal?

% focus on PERCEIVED mistakes -- this is why we omit robot speech

% and while we show that language can be used to identify some of the mistakes, sometimes it is not sufficient and that is why we look at facial expressions 

\section{Related Work}

As we aim to bridge the gap between research on affect recognition and adaptive human-robot dialogue, we begin by discussing related work in the theoretical foundations of affect, with a particular focus on emotion theories. We then explore recent advances in automatic affect recognition, and finally review related work in detecting mistakes in human-robot interaction.
%As we aim to bridge the gap between research on affect recognition and adaptive human-robot dialogue, we discuss first the theoretical concepts behind emotions and then related work in detecting mistakes in human-robot interaction.

\subsection{Confusion as an affective-cognitive state}
\textit{Affect} is a broad term that encompasses various constructs such as sentiment, feeling, mood, and emotion. These constructs are central to understanding human experience and behaviour, and they have been studied across multiple disciplines such as psychology, cognitive science, and affective computing. Among these, emotions have received particular attention due to their observable impact on decision-making, communication, and social interaction. Despite this focus, researchers have yet to converge on a single, universally accepted definition of what constitutes an emotion. Emotion research is typically categorized into three major theoretical frameworks: discrete or basic emotion theories~\cite{Plutchik1980, Ekman1992}, constructionist theories~\cite{Mehrabian1974, Posner2005}, and appraisal theories~\cite{Ellsworth1998, Scherer2005}. 

Discrete or basic emotion theories argue that humans have a set of biologically ingrained, universal emotions %, such as fear, anger or happiness, 
that are distinctly expressed through facial expressions. To apply this theory in data analysis, researchers use a fixed taxonomy of emotion labels to annotate texts, audio or videos~\cite{Mohammad2013, Livingstone2018, poria-etal-2019}.
Constructionist theories, on the other hand, suggest that emotions are learned constructs that are formed through a combination of core affective features (e.g., degree of pleasantness, level of arousal). For analytical purposes, emotions are measured in a constructionist way along the dimensions of arousal, valence, and dominance~\cite{busso2008iemocap, McKeown2012, buechel2017}.
Finally, appraisal theories focus on how we cognitively evaluate or ``appraise" salient stimuli along a range of checks (e.g., goal relevance, urgency, novelty). Although appraisals have recently found their introduction in the field of natural language processing (NLP), its application in the field of facial emotion recognition remains limited~\cite{Scherer2018, stranisci-etal-2022, troiano-etal-2023}.

%Even though many competing theories exist, there is some basic agreement on the fact that emotions are a dynamic processes that consists of different components: ~\cite{Scherer2022}.

Building on these theoretical foundations, it becomes evident that not all affective phenomena fit neatly within the boundaries of emotion as traditionally defined. One such phenomenon is \textit{confusion}, which we examine in this study as an affective-cognitive state that arises in natural, non-acted interactions. Confusion is often characterized by a blend of cognitive dissonance, uncertainty, and emotional arousal, making it a hybrid state that challenges the discrete categorization of emotions---though appraisal theories are capable of accounting for this nuanced state~\cite{Ellsworth2004}. It typically emerges in response to complex or ambiguous stimuli, particularly in learning or problem-solving contexts, where individuals struggle to reconcile conflicting information or expectations~\cite{DMello2014}. Rozin and Cohen~\cite{Rozin2003} observed that college students frequently found confusion, marked by eye-region facial movements, as a primary descriptor when interpreting facial expressions of emotion. Other researchers also remarked that the expression of confusion shares several characteristics commonly attributed to emotion, such as a consistent appraisal pattern and distinct facial cues~\cite{Craig2008,DMello2014}.

%Atapattu et al.~\cite{atapattu_identification_2019} describe confusion as both an emotional and cognitive state that emerges in unfamiliar situations or when one is confronted with complex information, such as during learning experiences. Additionally, research by Ketlner and Shiota~\cite{keltner_new_2003} highlights that confusion naturally occurs in response to stimuli that have insufficient or contradictory information. Although confusion is typically not included in standard taxonomies of basic emotions, Rozin and Cohen~\cite{Rozin2003} observed that college students frequently found confusion as a primary descriptor when identifying emotions expressed through facial expressions. 
%Leveraging video data as the primary input data enables the extraction of temporal features leading up to moments of confusion.

%related work on confusion specifically (see thesis + \url{https://psyche.co/ideas/perplexed-embrace-it-confusion-is-a-symptom-of-learning})

\subsection{Affect recognition using machine learning}
With advances in deep learning for computer vision, there has been a growing interest in automatically detecting emotional expressions from facial expressions, known as facial expression recognition (FER). Most FER research remains built on discrete emotion theories (i.e. classifying facial expressions into distinct classes as happy, sad, anger, surprise, fear, disgust, contempt, or neutral) \cite{alarcao_identifying_2018, zeng_face2exp_2022, ding_facenet2expnet_2016, githubFER}. FER systems are usually deep neural networks, built by taking models that are pre-trained for tasks such as detecting facial landmarks, and fine-tuning them on images of faces labelled for one of the basic emotions. Such models can achieve an accuracy of up to 96.8\% \cite{ding_facenet2expnet_2016}.

%Turning away from the specific confusion detection aspect, the more general field of facial expression recognition (\gls{FER}) is certainly a hot topic in the recent research, particularly for the classification of seven discrete emotions (happy, sad, anger, surprise, fear, disgust, contempt) and neutral \cite{alarcao_identifying_2018, zeng_face2exp_2022, ding_facenet2expnet_2016, githubFER}. Ding et al. describe that fine-tuning large pre-trained networks is an efficient approach to perfect facial expression recognition \cite{ding_facenet2expnet_2016}. They do this in two stages, in which the first one leverages the features from an already pre-trained face net. Next, the pre-trained network is fine-tuned by using labelled data.

However, such models are trained and evaluated in a rather black-and-white way: either the expression is indicative of one of the basic emotions, or it is neutral.
%either one of the basic emotions is present, or the expression is neutral.
This optimises them to learn extreme, prototypical expressions of these basic emotions \cite{ding_facenet2expnet_2016}. However, as Kappas et al. note \cite{kappas2020communicating}, such ``full-blown patterns'' are rarely present in the real world, and they are not a reliable indicator of emotions. This might limit the applicability of facial expression recognition models in real-world settings, such as human-robot interaction.

%(This line was in the HRI paper)
%Zhang et al. argue that facial expression cannot be captured in just seven basic emotions and opt to use open-set samples \cite{zhang_open-set_2024}. Open-set samples are particularly interesting because this research tries to move away from a set of predefined emotions that are unable to capture every single emotional intricacy, by introducing initially undefined classes. Other methods include dimensional emotion models that capture affective features such as valence and arousal\cite{alarcao_identifying_2018, mertens_findingemo_2024}. By using these techniques, researchers are less restricted by certain categories, making it easier to represent emotional tendencies without a prior label. 

%Arousal corresponds to the intensity of an emotion, while valence captures the positive or negative connotation \cite{citron_emotional_2014}. 
%An emotion label tries to capture the whole context instead of being purely based on a specific expression.
%Levering these metrics could ease the path towards the classification of more niche emotions such as confusion and especially the reaction to a robot mistake.

% paper(s) Hatice
% maybe papers Arvid Kappas

\subsection{Mistake detection in human-robot interaction}

% The automatic detection of confusion through facial expressions could enhance applications in various fields, such as education \cite{atapattu_identification_2019}, customer service \cite{banga_emotion_2023}, and healthcare \cite{postma-nilsenova_automatic_2015}, serving as an early indicator for identifying potential robot mistakes.
% By proactively identifying these mistakes, the overall efficiency of instructions significantly increases. 

% Instead of purely conversational research, many researchers combine this with a certain task that has to be executed correctly by the participant \cite{reciperobot, headandshoulder, toerrisrobot}. 
% It is quite common in this field to use the Wizard-of-Oz approach to deliberately add mistakes to the conversation.
% In addition to that, there exists a field of research focussed on robot mistakes in a non-social context, that occur purely during the execution of a specific task from the robot \cite{stiber2022modeling}.

Studying user responses to robot failures and automatically detecting them has been a topic of research in human-robot interaction for at least a decade \cite{Kontogiorgos2021ASC}. 
Most of the research in this direction focuses on some type of collaborative robots, where the user tries to accomplish a task together with the robot or with the robot's guidance.

For example, Trung et al. \cite{headandshoulder} investigate automatically detecting a Nao robot's failures while it is giving instructions to a user who is completing a LEGO construction task. By examining only the user's head and shoulder movements, their model can detect robot technical failures. However, it is less effective when the robot violates social norms or when it encounters unfamiliar users.
Stiber et al. \cite{stiber2022modeling, stiber2023using} also look at a collaborative robot: their study employs a robot arm that picks up objects in response to user commands. They capture users' implicit feedback through facial expressions by employing a pre-trained model that detects Facial Action Units (FAUs), which are specific movements in the face. A neural network then classifies these facial movements to determine whether they are indicative of robot failure. 

Kontogiorgos et al. \cite{kontogiorgos2020behavioural} identify conversational failings of a Furhat robot during an assembly task by analyzing head movements, gaze, and speech signals. They find that external annotators can most reliably detect robot mistakes by identifying when the user is in a state of \textit{confusion}.
Comparing this task with another context, where the user and robot engage in a negotiation task \cite{Kontogiorgos2021ASC}, they suggest that non-verbal features indicating affect and emotion are more generalisable across interaction contexts, whereas verbal feedback is highly context-specific.

% compared multiple 
% (physical collaboration , no facial expressions except for showing that "confusion" is the most reliable emotion through which external annotators can detect robot failures, \cite{Kontogiorgos2021ASC} comparing with negotiation (verbal-only), indicated that non-verbal features indicating affect and emotion seem to be more generalisable across interaction contexts), while verbal feedback is highly biased towards the interaction context

Perhaps most similarly to our study, Axelsson et al. \cite{axelsson2022multimodal} investigate multimodal user feedback to a robot that presents paintings, with external annotators labelling the user's feedback as positive, negative, or neutral. The robot, controlled through a Wizard-of-Oz setup, would occasionally misspeak (replacing part of the robot's speech by other words or silence). The authors find that facial expressions, backchannels, and speech tone are most indicative of negative feedback, and used Random Forests and LSTMs to automatically classify the feedback. In that study, facial expressions were manually annotated for specific movements such as frowning. In a follow-up study with an autonomous, real-time robot, the authors did not use automatic facial expression detection, instead using only head movements and speech features \cite{axelsson2023you}.

%(did look at facial expressions and other modalities in a more social, educational interaction, only study in social interaction, also used an LSTM to process temporal dimension) (-> just trying to replicate this with different error types?)

Notably, none of these studies collect explicit feedback from the user as ground truth for training or evaluation. Rather, they solely rely on external annotations or explicitly introduced failures and do not examine whether the user perceived them as failures---Trung et al. \cite{headandshoulder} have already found some cases where users do not perceive the robot's designed social norm violations as such.

Our study builds upon previous research by focussing on facial expressions, which have been shown to be a crucial feedback modality. We investigate this within a social, dialogue-only human-robot interaction, rather than a physical collaboration task, examining four distinct failure types and uniquely collecting ground truth from the users.

\section{Datasets}

Two datasets are used in this research: a dataset of natural user feedback in human-robot dialogues and a dataset containing videos with emotionally expressive content, sourced from an online repository.

\subsection{Human-robot conversation dataset}
The REPAIR-Corpus dataset is an in-house corpus that captures the natural reactions and social cues of participants during human-robot interactions. It is available through Zenodo~\cite{furhat_cooking_data}. %\footnote{The dataset is anonymously submitted to HRI as a separate paper and will be released after review.}
The dataset consists of 240 videos featuring 40 participants, each of whom engaged in six conversations with the social robot Furhat \cite{al2012furhat}.
%\footnote{\url{https://furhatrobotics.com/}}.
The videos show the user from the robot's perspective and were recorded using an external webcam placed in front of the robot.
Participants were told that they would be interacting with six different systems, while the conversations were in reality fully scripted and grounded in cooking recipes, with each recipe representing a different ``system''. The scripts incorporated four types of robot mistakes: interruptions, oversharing information, undersharing information, and irrelevant comments~\cite{Torrey2006, Chakraborti2017, Serholt2020}. The order of these mistakes was randomized across the different recipes. 
Additionally, the order in which participants encountered the recipes was randomized. To address these communication errors, the experiment introduced repair strategies that differed in formality (formal vs. informal), including apologies, promises, and explanations~\cite{Esterwood2023}. However, participants often found these strategies to be somewhat unnatural, especially when they disrupted the flow of the conversation.

During the conversations, participants were asked to press a binary controller button after their reply to each robot utterance to indicate whether they perceived the utterance as a mistake or inappropriate for flow of the conversation.
%In our experiments, we use this feedback as a proxy for participants experiencing confusion. We recognize that this heuristic simplifies the complexity of the real data, as participants' perceptions of mistakes are highly subjective and their replies may reflect other emotional states beyond confusion. Nevertheless, after a closer manual review, we believe there are enough natural expressions of confusion to justify this research approach. Furthermore,
In our experiments, we use this feedback as the ground truth for whether the user experienced a miscommunication due to a robot mistake.
The dataset includes a substantial amount of data capturing emotional expressions, thus distinguishing itself from many other existing datasets that rely on acted emotion expressions~\cite{busso2008iemocap, Goodfellow2015, poria-etal-2019}, which are often used in facial expression recognition research.

%\paragraph{Preprocessing of the data.} 

%Maybe not a subsubsection?
% Cropping videos and cutting them into conversational turns -> do we want to include this?
%Using this dataset requires some preprocessing. The first step is to match the timing of the controller logs with the actual videos.
%Next, the lengthy videos were cut into separate fragments containing a robot utterance and a user's response based on the dialogue flow. Following the cutting procedure, the videos are cropped to display the facial expressions more clearly.

%Data splits

To use this dataset in our experiments, we first process the dataset into short video fragments which each contain one exchange, consisting of a robot utterance and user response. This collection of 2600 fragments, 27.2\% of which were labelled by the user as containing a mistake, was further divided into a training, validation, and test set, which respectively contain 67.5\%, 22.5\%, and 10\% of the samples. In each subset, labels follow the same distribution as in the complete dataset, and each participants' samples are contained within the same subset.

%To train a model on this dataset, we created train, validation and test splits, taking into account the stratified and grouped aspect of the data. The test set contains 10\% of the dataset, and the remaining 90\% is assigned to the train-validation set. This is further divided into the train set (75\% of the train-validation set) and the validation set (25\% of the train-validation set).

%Imbalanced data -> what was used in the final model?

\subsection{Expressive confusion dataset}

To validate the performance of our mistake detection model and as existing datasets either do not label confusion or do not contain video data, we constructed a small dataset that consists of short videos showing unambiguous expressions of confusion.
As the videos contain clear expressions of confusion, we can investigate whether our model is capable of recognizing a confused state from facial expressions.
Furthermore, the videos are short, so the problem of extracting salient moments showing emotional expressions does not impact performance on this task.
%it is the users' manner of expressing that they are perceiving a robot mistake, which forms the bottleneck, or whether our model is not able to recognise emotions in facial expressions.
139 videos were collected from the public repository Tenor\footnote{https://tenor.com/en-GB/}, selecting videos that contained either an expression of confusion or a neutral expression. All included videos contain only one person with their face clearly visible, who is exhibiting at least some movement, and appears to be in an interaction with another person. The videos also do not show any visual effects and the camera remains (mostly) stationary. All videos have a frame rate of 10 frames per second, reducing the complexity of the data, and are at most four seconds long. The dataset is imbalanced on purpose (30\% neutral, 70\% confusion) to resemble the imbalanced distribution of the REPAIR-Corpus dataset, although the imbalance is in the opposite direction, focusing on training the model to detect confusion. The dataset is split into a training and test set with an 80-20 split.

% % STILL TO BE REWRITTEN

% Due to the low accuracy of the models on the Recipe dataset, a small, more expressive video dataset was introduced as a sanity test. %This wording might be weird 
% The result is a rather small dataset of short videos with expressive and unambiguous emotions, collected from the Tenor website \cite{tenor}. It allows for an initial understanding of models capable of working with temporal data. Furthermore, creating models with acceptable performance on the GIF dataset, proves that when using an unambiguous dataset, a model is capable of making a distinction between confused and neutral expressions. The videos of neutral expressions were chosen in such a way that they did not only capture stationary moments, preventing the model from solely focusing on the detection of movement to define to which class an expression belongs. The expressive dataset was made to be imbalanced (30\% neutral, 70\% confusion) to resemble the imbalanced distribution of the RoboRecipeMistakes dataset, although the imbalance is in the opposite direction.

\section{Miscommunication detection system}

In this section, we describe and evaluate our miscommunication detection system.
The system contains two main components. First, the input -- a video that shows one human-robot exchange -- is cropped to contain only the user's face, and then given to an algorithm which extracts the most salient moments from this video. These moments are then passed on to the second component, which is a model that classifies the exchange as containing a miscommunication or not. Both components leverage high-level features which are extracted from the videos using existing off-the-shelf models.
We first describe these feature extraction models and then discuss and evaluate the salient moment extraction algorithms.
Then, we present the miscommunication classification model, first evaluating it on the human-robot dialogue ``REPAIR-Corpus'' dataset and then investigating how it performs on short videos containing more pronounced emotional expressions.

%We first present the construction and evaluation of our system using the real-world RoboRecipeMistakes dataset, separating the relevant extraction algorithm from the classification model. Then, we apply our model to the expressive dataset, investigating how it performs on videos containing much more expressive emotions. Finally, we report how humans perform at this task, and explore what the role of foundation models can be.

\subsection{Visual feature extraction}
\label{sec:backbones}

Since raw videos are composed of numerous consecutive still images, their high dimensionality makes them unsuitable for direct input into our model. Therefore, we leverage pre-existing models to extract lower-dimensional representations of the images, capturing essential high-level visual features. We compare three approaches for this, of which the latter two are specialised for facial expressions.

First, we look at general-purpose computer vision neural networks that are originally trained for image classification. Specifically, we compare the convolutional neural network VGG16 \cite{simonyan2014very} and the residual neural network ResNet50 \cite{he2016deep}. For both models, the final classification layer is removed, so we use the output of the final hidden layer.

Second, we use a model that extracts facial keypoints, provided by Google's MediaPipe framework \cite{lugaresi2019mediapipe}. It detects the location of 468 keypoints on a human face, focusing on more expressive parts of the face.

Third, we use a different model from MediaPipe which extracts \textit{blendshapes} instead. Popular in the domain of digital avatars, this approach extracts 52 scalar values that together control a 3D mesh model of a human face, and is similar to the detection of FAUs, which are often used in related work.

%Finally, we use a model that directly predicts emotions from 

\subsection{Salient moment extraction}

\begin{figure}
    \centering
    \includegraphics[width=\columnwidth]{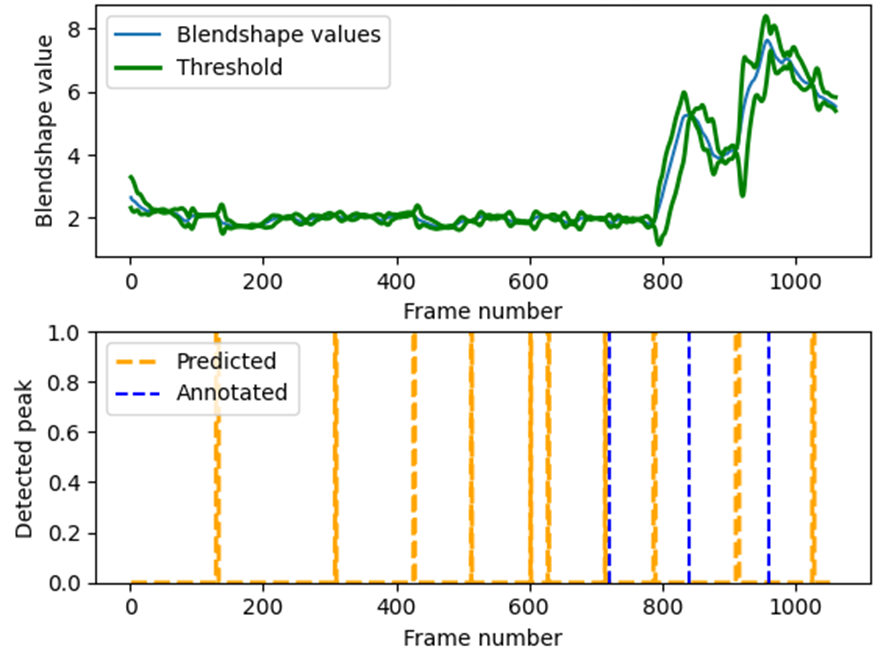}
    \caption{Illustration of the real-time peak detection algorithm for extracting salient moments, applied on one video fragment. The top graph shows the sum-of-blendshapes signal and the threshold calculated by the algorithm. The bottom graph shows which salient moments the algorithm predicted in blue, with the manually annotated moments indicated in orange.}
    \label{fig:peak-blend}
\end{figure}

As the video fragments in the dataset are too large to be efficiently used as input for a model, and as large parts of the videos often contain no movement, we attempt to extract the most salient moments for mistake detection.
As a heuristic, we consider any moment where there is significant movement in the user's facial expression as a salient moment.
We compare three approaches to detect these moments: one that is based on extracted blendshapes and the other two use facial keypoints.

The approaches are evaluated on a subset of 20 video fragments from the REPAIR-Corpus dataset. For each fragment, salient moments were manually annotated, for a total of 49 salient moments in the 20 fragments. Each salient moment is identified by a single frame in the fragment. Predicted salient frames are counted as correct when they are less than 60 frames (one second) removed from an annotated salient frame.

In the blendshape approach, each of the 52 blendshape values is first smoothed through a linear convolution over the length of the video fragment, with a window size of 45 frames. This reduces the influence of local extrema, increasing the robustness of the approach.
However, taking these 52 separate values into account would be overly complex, as the occurrence of salient moments is not easily mapped to specific behaviours of the separate blendshapes, and applying the same algorithm to all blendshapes at once would result in the entire video being seen as salient. Therefore, all 52 values are summed into one combined scalar signal. 
This combined signal is then passed onto a ``real-time'' peak detection algorithm adapted from \cite{brakel2014}. For every frame, the algorithm calculates a threshold based on the values of the signal in the previous frames. If the signal exceeds that threshold, the frame is identified as salient. The working of this algorithm is illustrated in Figure~\ref{fig:peak-blend}.
After tuning the parameters of the peak detection algorithm, this approach is able to extract 24 of the 49 annotated salient moments, or 49\%.

Second, we evaluate an approach that uses extracted facial keypoints. As each keypoint is characterised by $(x,y)$ coordinates in the image, we consider the sum, over all keypoints, of the square of the distance to the keypoint in the previous frame, again creating a single scalar signal.
After smoothing this signal with a linear convolution with a window size of 45 frames, it is passed to the same real-time peak detection algorithm as in the previous approach. This approach is able to detect 17 out of the 49 keypoints, or 34\%.

Finally, we improve the keypoint-based approach by using a different algorithm, that instead selects the three frames that have the highest sum-of-square-distances with the previous frame, selecting only frames that are at least 60 frames removed from each other. This approach extracts 30 of the 49 annotated salient moments, or 61\%. As it performed best out of the three tested salient moment extraction approaches, we use this approach for our future evaluations.

\subsection{Classification model}

% Architecture(s)

% Evaluation

% (Wouldn't it fit best to put the results here?)

In the previous section, we developed a salient moment extraction algorithm. As this gives us the three most salient moments per video fragment, we extract the 60 frames around each moment, and concatenate these three mini-clips. The resulting video is a three-second compilation of the most salient fragments from the original video fragment, and can be used as input to our model.

Our model is constructed using the following basic architecture. First, one of the visual feature extraction models described in Section~\ref{sec:backbones} processes each frame of the video into a lower-dimensional representation. The resulting sequence is used as input for an LSTM neural network, which is designed to process temporal data. A final linear layer reduces the output of the LSTM to one scalar value, which is used to classify whether a mistake was present in the input video.

We evaluate three visual feature extraction models: VGG16, ResNet50, and MediaPipe's blendshapes. In addition, we explore three strategies to enhance model performance: applying a weighted loss function to address dataset imbalance, splitting the samples so that each relevant moment is treated as a separate sample, and shortening the video length by keeping only every $n$-th frame, comparing different values of $n$. For each approach, we tune the hidden size of the LSTM, the batch size, and learning rate of the model.

%We report the model's performance using the F1 score on the validation set, since the dataset is imbalanced. As only 27.2\% of the samples in the dataset contain robot mistakes, a dummy baseline that always predicts the presence of a mistake would achieve an F1 score of 0.43\%.
We report the models' performance by showing their ROC curve, following the example set by prior work \cite{headandshoulder, kontogiorgos2020behavioural}. To quantify that performance, we calculate the area under the ROC curve (AUC). An AUC of 1.0 represents a perfect classifier, 0.0 a classifier that is always wrong, and 0.5 a random classifier.
Trung et al. reported AUC scores of up to 0.66 \cite{headandshoulder}, while the ROC curves reported by Kontogiorgios et al. showed even better performance without providing the exact AUC \cite{headandshoulder,kontogiorgos2020behavioural}.
%Prior work on detecting mistakes in human-robot interactions reported AUC scores of 
%That prior work reported AUC scores of 

\begin{figure}
    \centering
    \includegraphics[width=0.9\linewidth]{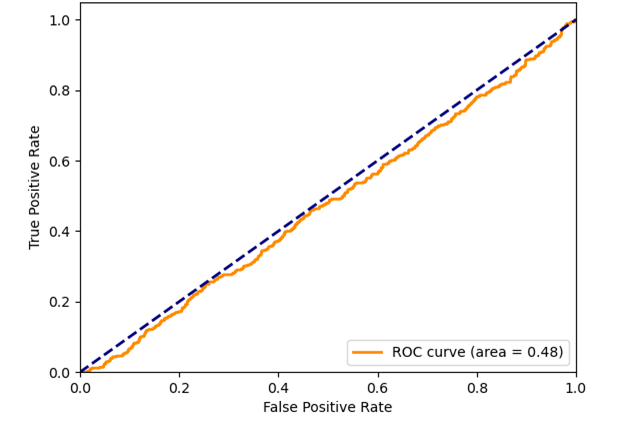}
    \caption{ROC curve of the best-performing miscommunication detection model on the REPAIR-Corpus dataset.}
    \label{fig:recipe-learningcurve}
\end{figure}

However, not a single one of the tested model configurations performs better than a random classifier.
%achieved a score that is higher than this baseline.
The best-performing model is the ResNet-based model, whose ROC curve is shown in Figure~\ref{fig:recipe-learningcurve}. It obtains an AUC of 0.48, which is slightly worse than random chance.
% In Figure~\ref{fig:recipe-learningcurve}, we show the %training curves
% ROC curve of the model that achieved the best performanc, with
% %highest validation F1 score
In this model, the samples are split up into single salient moments and only every fifth frame is retained. The model was trained for 50 epochs, with the following optimal hyperparameters: batch size of 16, learning rate of $1e^{-4}$ and an LSTM input size of 52 and hidden size of 256. The training and validation accuracy curves show that the model quickly overfits on the training set, implying that longer training would not improve performance on the validation or test set. 
%On the test data, this model achieved an F1 score of
%With an area under the ROC curve of 0.48, it actually performs slightly worse than random chance.

% \begin{figure}
%     \centering
    
%     \begin{subfigure}{0.4\textwidth}
%         \includegraphics[width=\columnwidth]{figures/recipe-learningcurve.png}
%     \caption{Training curves.}
%     \end{subfigure}
%     \begin{subfigure}{0.4\textwidth}
%         \includegraphics[width=\columnwidth]{figures/recipe-roc.png}
%     \caption{ROC curve.}
%     \end{subfigure}
%     \caption{Training curves and ROC curve of the best-performing mistake detection model on the RoboRecipeMistakes dataset.}
%     \label{fig:recipe-learningcurve}
% \end{figure}

\subsection{Performance on expressive confusion dataset}

\begin{figure}
    \centering
    \includegraphics[width=0.9\linewidth]{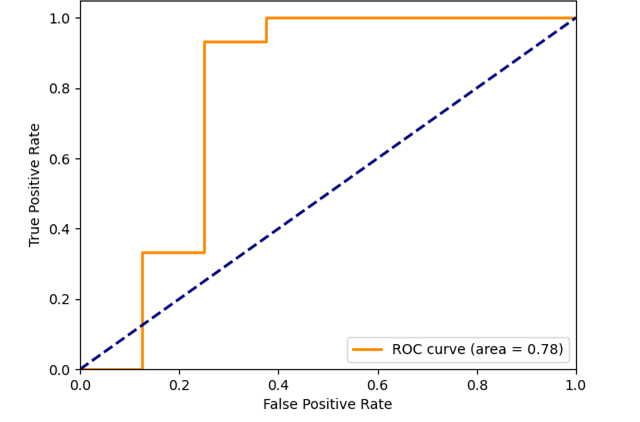}
    \caption{ROC curve of the best-performing miscommunication detection model on the expressive confusion dataset.}
    \label{fig:gif-dataset}
\end{figure}

Our mistake detection system performed worse than expected on the human-robot conversation dataset, underperforming prior research. This leaves a question to be answered: is this due to our methodology, meaning our model or training approach, or because of the data? In other words, is the model we designed capable of detecting facial expressions that indicate a user experiencing a miscommunication?
%able to detect emotions?

Therefore, as a toy problem to show the suitableness of our model, we apply the same methodology used for our classification model to our dataset of short videos with unambiguous expressions of confusion. This was identified by prior work to be the most reliable expression indicating miscommunications \cite{kontogiorgos2020behavioural}.
%, which is smaller than the RoboRecipeMistake dataset.

%We use the same model architecture as for the previous dataset.
%Opting to use the blendshapes as visual feature extraction because of its lower dimensionality (which will likely work better given the limited size of this dataset), 
Using the same model architecture as for the previous dataset, the results show that our approach is able to recognise confusion in facial expressions.
As shown in Figure~\ref{fig:gif-dataset}, the area under the ROC curve is 0.78, indicating that the model clearly exceeds random chance.
The model reaches an F1 score of 74.1\% and accuracy of 69.7\%. Learning curves also show the model is not overfitting.
%Observing the learning curves show the model is not overfitting, and as shown in Figure~\ref{fig:gif-dataset}, the area under the ROC curve is 0.78, showing that the model is clearly exceeding random chance, which is also confirmed by the confusion matrix for the validation data.
These results were achieved when using blendshapes as visual features extractor. The model was trained for 75 epochs, with a learning rate of $1e^{-4}$, batch size 8, and an LSTM with input size 52 and hidden size 512. No frames were dropped, as the samples were only 40 frames long.

% \begin{figure}
%     \centering
%     \begin{subfigure}{0.4\textwidth}
%         \includegraphics[width=\columnwidth]{figures/gif-curve.png}
%         \caption{Training curves.\\}
%     \end{subfigure}
    
%     % \begin{subfigure}{0.4\textwidth}
%     %     \includegraphics[width=0.9\columnwidth]{figures/gif-cm.png}
%     %     \caption{Confusion matrix.\\}
%     % \end{subfigure}
    
%     \begin{subfigure}{0.4\textwidth}
%         \includegraphics[width=\columnwidth]{figures/gif-roc.png}
%         \caption{ROC curve.}
%     \end{subfigure}
%     \caption{Training and ROC curves of the best mistake detection model on the expressive confusion dataset.}
%     \label{fig:gif-dataset}
% \end{figure}

% hier gpt-4o op gif dataset?

\section{Human evaluation}

Up to this point, we have demonstrated that our mistake detection system, despite being effective at detecting confusion from facial expressions, performed no better than random chance on a dataset of real human-robot conversations. Now, we aim to explore the level of performance we can reasonably expect on this dataset. How accurately can humans identify user reactions to robot miscommunications?

A human evaluation was set up to answer this question, recruiting 17 annotators who were not familiar with the original study.
Each annotator was shown the same 20 video fragments sampled from the REPAIR-Corpus dataset, each containing one exchange between the robot and the user. This sample reflects the distribution of the full dataset: 6 of the fragments were labelled as a miscommunication by the user's button feedback, the remaining 14 were not. 

Whole fragments were shown to the annotators, meaning the fragments were not passed through the salient moment extraction algorithm and not cropped to the user's face. No audio was included, neither robot speech nor user speech, to completely match the information available to our model and to eliminate any bias from the content of the robot's speech -- the aim is for raters to label whether the participant of the original experiment perceived the robot speech as a miscommunication, not whether they themselves perceive it as such.

%As the miscommunications in the dialogues are scripted, they can indicative of whether the robot was intended to make a mistake. We want raters to label whether the user perceived the robot speech as a mistake, not whether they themselves perceive the robot speech as a mistake.

Participants were asked to label each video one by one as ``Mistake'' or ``No mistake'', being instructed to label videos as ``Mistake'' whenever they thought the robot had made a mistake, the participant was confused, the interaction strayed off its normal course, or it seemed the robot said something strange or wrong.

%The results are sobering.
%Looking at each video separately, as shown in Table~\ref{tab:human}, we see that on average, a video containing a mistake is correctly identified by 50\% of the participants. Moreover, more than half of the videos containing a mistake, are misclassified by a majority of participants. Videos with no robot mistake are, on average, correctly identified by 59\% of the participants. While some of the videos without mistake reach high human agreement, these videos tend to show a lack of any response by the user.

\begin{figure}
    \centering
    \includegraphics[width=.9\columnwidth]{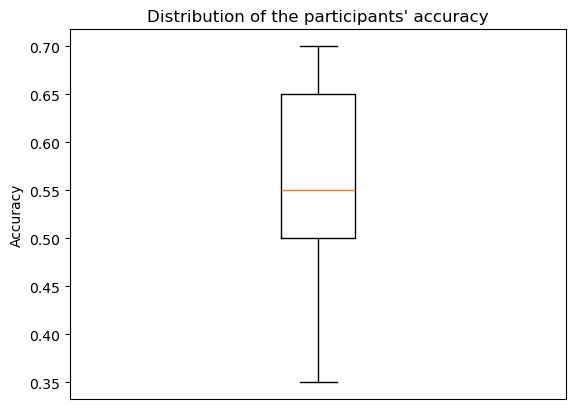}
    \caption{Distribution of the participants' ($n=17$) accuracy at identifying videos with robot mistakes.}
    \label{fig:human-acc}
\end{figure}

Results show that human raters perform no better than our classification model. On average, participants correctly classified 56\% of the fragments. Performance is also widely spread across participants: as shown in Figure~\ref{fig:human-acc}, the lowest accuracy of any participant was 35\%, with the highest 70\%.
Calculating the inter-annotator agreement (IAA) using Fleiss' Kappa~\cite{LandisKoch1977}, which ranges from -1 to 1, returns a value of -0.012, meaning absolutely no agreement between the raters.

Looking at performance for each video fragment separately, we see that on average, a video containing a miscommunication is correctly identified by 50\% of the participants. More than half of the videos containing a miscommunication are misclassified by a majority of participants. Videos without a miscommunication are, on average, correctly identified by 59\% of the participants. While some of the videos without miscommunication reach high human agreement, these videos tend to show a neutral facial expression.

%These results show that the human annotators performed no better than our classification model.

% \begin{table}[]
%     \centering
%     \caption{Percentage of Participants ($n=17$) Correctly Classifying Each Video}
%     \begin{tabular}{lcc}
        
%         \textbf{Class} & \textbf{Video} & \textbf{Percentage of participants} \\ \hline 
%         \multirow{7}{*}{Miscommunication} & 0 & 76.47\% \\  
%          & 3 & 35.29\% \\ 
%          & 4 & 41.18\% \\ 
%          & 8 & 41.18\% \\ 
%          & 9 & 76.47\% \\ 
%          & 14 & 29.41\% \\ 
%          & \textbf{Average} & \textbf{50.00\%} \\ \\
%         \multirow{15}{*}{No miscommunication} & 1 & 82.35\% \\  
%          & 2 & 82.35\% \\  
%          & 5 & 64.71\% \\ 
%          & 6 & 35.29\% \\ 
%          & 7 & 47.06\% \\ 
%          & 10 & 47.06\% \\ 
%          & 11 & 88.24\% \\ 
%          & 12 & 58.82\% \\ 
%          & 13 & 58.82\% \\ 
%          & 15 & 52.94\% \\ 
%          & 16 & 58.82\% \\ 
%          & 17 & 29.41\% \\ 
%          & 18 & 82.35\% \\ 
%          & 19 & 41.18\% \\ 
%          & \textbf{Average} & \textbf{59.24\%} \\ 
%         \end{tabular}
    
%     \label{tab:human}
% \end{table}

%Second, we compare the participants' agreement and performance over all videos.
%In addition, when looking at the distribution of all participants' accuracy over all 20 videos, we see accuracies range from 35\% to 70\% with a mean accuracy of 56\%, as shown in Figure~\ref{fig:human-acc}.
These results clearly indicate that humans are not able to reliably identify miscommunications from only users' facial expressions and body language in these real-life human-robot conversations.

% muted audio so that we are not biased by the scripted nature of the dataset, and not biased by our third-party expectation of whether we think the robot made a mistake or not

% hier het foundation model ding op het verbale aspect?

% maybe only verbal aspect through foundation model?

\section{Discussion and Conclusion}

This work investigates the performance of a system that detects miscommunications in social human-robot interactions from users' facial expressions. Prior work has identified facial expressions as an important modality for users to express feedback to robots, but does not yet automatically detect facial expressions \cite{axelsson2022multimodal, kontogiorgos2020behavioural} or focuses on physical human-robot collaborations rather than social dialogue-based interactions \cite{stiber2023using}. Furthermore, prior work relies on external annotators to label interactions as miscommunications \cite{axelsson2022modeling} or uses purposefully designed robot failures without verifying that users perceive them as such \cite{Kontogiorgos2021ASC}. This notion is particularly uncertain in the context of social failures \cite{headandshoulder}.
% related work: first to use facial expressions automatically in social HRI, prior work identified facial expressions and emotional signals as more promising
% first to user user-labelled miscommunications
% prior work showed that humans should be able to identify miscommunications (dimos: ), agnes:

Aiming to bridge the gap between stellar results in facial expression recognition and human-robot interaction, we built a system that processes robot-perspective videos of human-robot exchanges and detects whether they contain a robot-caused miscommunication, examining the user's facial expressions.
This system first extracts the most salient moments in the video before classifying it. Given the lack of facial expression recognition models capable of processing videos and identifying complex affective-cognitive states such as confusion, crucial for detecting miscommunications, our system uses pre-trained models that track facial movements.%facial keypoints or blendshapes.

We trained and evaluated this system on REPAIR-Corpus: a multimodal dataset of educational human-robot conversations. While the robot was explaining recipes to the user, it introduced four types of scripted miscommunications. After each robot utterance, the user pressed a button to indicate whether they experienced this as a miscommunication.

However, our system performs below expectations. Despite evaluating multiple configurations, it never performs better than a random classifier. Investigating the cause of this low performance, we first tested our system on a toy dataset which is more similar to traditional facial expression recognition datasets, containing very clear expressions of confusion. The system performed well on this dataset, showing the suitability of our technical approach.

We then ran a human evaluation study to assess whether external annotators are able to identify miscommunications in our dataset. While prior work had shown reasonable success in detecting miscommunications from videos without audio \cite{kontogiorgos2020behavioural, axelsson2022modeling}, our participants showed absolutely no agreement when classifying videos from our dataset.

These results highlight an underexposed question in human-robot interaction: when do users genuinely intend to convey (negative) feedback to a robot? Our results indicate a discrepancy between the occurrence of miscommunications, even as perceived by users, and the moments when users express them in a way that is noticeable even to other humans. Combined with the challenge that feedback is swift and much more subtle than can be classified into basic emotions, it can explain why automatic detection of miscommunications or mistakes performs worse than expected in social human-robot interactions.

Other questions are raised as well. Do people convey miscommunications through the same facial expressions and to the same extent to robots as to other humans? Would the human annotators do better at recognising miscommunications in human-human interactions?
Additionally, we know that context is very important when interpreting facial expressions. % \cite{hoegen2019signals}.
What is the impact of different feedback modalities, such as the user and robot utterances, on recognition performance?
Finally, we recognise that our study took place in a lab setting and cannot fully approximate a real-world interaction. Participants likely expected robot mistakes, and the miscommunications did not have any immediate consequences to them.
Could the presence of a feedback button have impacted the users' tendency to convey non-verbal (or verbal) feedback?
Or did the robot's lack of reaction to their feedback have an impact?

% do people show same facial expressions with robots as with people?
% are people better at recognising these occurences in human-human interactions?
% does the inclusion of other modalities provide better results? context is important, especially for states such as confusion rather than basic emotions with consistent facial expression mappings
% how did the presence of the feedback button impact this, e.g. did they expect the presence of mistakes? did they still feel the need to convey feedback through facial expressions, when they has the button, and notice that the robot doesn't react to their expressions
% not a real-world scenario, rather is lab setting, no immediate consequences for participant

Future work should explore the factors that affect when users intend to communicate their feedback in a perceivable manner, such as the interaction context, the robot's embodiment or perceived autonomy, and other potential influences. Additionally, it is crucial to integrate these findings with the dialogue context, as well as verbal and other non-verbal feedback.
These insights will aid in designing better human-robot interactions by highlighting the complexity of user feedback in social human-robot conversations. They also present clear challenges to facial expression recognition research regarding its applicability in autonomous human-robot interaction.

\balance

\bibliographystyle{IEEEtran}
\bibliography{lib}

\end{document}